\documentclass[conference]{IEEEtran}
\IEEEoverridecommandlockouts
\usepackage{orcidlink}
\usepackage{cite}
\usepackage{amsmath,amssymb,amsfonts}
\usepackage{algorithmic}
\usepackage{graphicx}
\usepackage{textcomp}
\usepackage{xcolor}
\usepackage{graphics}
\usepackage{multirow}
\usepackage{diagbox}
\usepackage[capitalise]{cleveref}  
\usepackage{subcaption}     
\usepackage{paralist}       
\usepackage{pdfpages}       
\def\BibTeX{{\rm B\kern-.05em{\sc i\kern-.025em b}\kern-.08em
    T\kern-.1667em\lower.7ex\hbox{E}\kern-.125emX}}
\begin{document}

\title{
PIM: Physics-Informed Multi-task Pre-training for Improving Inertial Sensor-Based Human Activity Recognition
}

\author{
\IEEEauthorblockN{Dominique Nshimyimana \orcidlink{0009-0009-8580-1248}}
\IEEEauthorblockA{
\textit{RPTU Kaiserslautern-Landau}\\
and
\textit{DFKI}\\
Kaiserslautern, Germany \\
}
\\
\IEEEauthorblockN{Bo Zhou}
\IEEEauthorblockA{
\textit{RPTU Kaiserslautern-Landau}\\
and 
\textit{DFKI}\\
Kaiserslautern, Germany}
\and
\IEEEauthorblockN{Vitor Fortes Rey \orcidlink{0000-0002-8371-2921}}
\IEEEauthorblockA{
\textit{RPTU Kaiserslautern-Landau}\\
and 
\textit{DFKI}\\
Kaiserslautern, Germany}
\and
\IEEEauthorblockN{Sungho Suh}
\IEEEauthorblockA{
\textit{RPTU Kaiserslautern-Landau}\\
and 
\textit{DFKI}\\
Kaiserslautern, Germany}
\\
\IEEEauthorblockN{Paul Lukowicz}
\IEEEauthorblockA{
\textit{RPTU Kaiserslautern-Landau}\\
and 
\textit{DFKI}\\
Kaiserslautern, Germany}
}

\maketitle

\begin{abstract}
Human activity recognition (HAR) with deep learning models relies on large amounts of labeled data, often challenging to obtain due to associated cost, time, and labor. Self-supervised learning (SSL) has emerged as an effective approach to leverage unlabeled data through pretext tasks, such as masked reconstruction and multitask learning with signal processing-based data augmentations, to pre-train encoder models. However, such methods are often derived from computer vision approaches that disregard physical mechanisms and constraints that govern wearable sensor data and the phenomena they reflect. In this paper, we propose a physics-informed multi-task pre-training (PIM) framework for IMU-based HAR. PIM generates pre-text tasks based on the understanding of basic physical aspects of human motion:  including movement speed, angles of movement, and symmetry between sensor placements. Given a sensor signal, we calculate corresponding features using physics-based equations and use them as pretext tasks for SSL. This enables the model to capture fundamental physical characteristics of human activities, which is especially relevant for multi-sensor systems. Experimental evaluations on four HAR benchmark datasets demonstrate that the proposed method outperforms existing state-of-the-art methods, including data augmentation and masked reconstruction, in terms of accuracy and F1 score. We have observed gains of almost 10\% in macro f1 score and accuracy with only 2 to 8 labeled examples per class and up to 3\% when there is no reduction in the amount of training data.
\end{abstract}

\begin{IEEEkeywords}
Human activity recognition, Self-supervised learning, Physics-inspired neural networks, Wearable sensors, Pretext tasks
\end{IEEEkeywords}

\section{Introduction}
Wearable sensor-based human activity recognition (HAR) has become increasingly significant due to its wide range of applications in fields such as healthcare \cite{zheng2017design, xu2018geometrical}, sports performance analysis \cite{zhou2022quali,singh2024novel}, smart home automation \cite{bianchi2019iot}, and operators' safety in manufacturing \cite{tao2018worker,suh2023worker,bello2024tsak}. Wearable HAR systems utilize data from devices equipped with sensors like accelerometers and gyroscopes, typically placed on the human body to detect and classify activities such as walking, running, or sitting. These systems are essential in personalized health monitoring, rehabilitation, and even elder care, where accurate recognition of daily activities can provide valuable insights into an individual’s physical condition and lifestyle.

Despite these promising applications, wearable sensor-based HAR faces considerable challenges, one of the most prominent being the collection and labeling of large-scale datasets. Annotating sensor data is often time-consuming, costly, and subject to privacy concerns, as data collection requires consistent user participation and manual labeling. This lack of large amounts of labeled data impacts the development of accurate and reliable deep learning-based HAR models, limiting their applicability in real-world scenarios. 

Self-supervised learning (SSL) has emerged as a viable solution to address the data scarcity issue in HAR by enabling models to learn from unlabeled data through solving pretext tasks \cite{haresamudram2022assessing}. Existing SSL methods for HAR commonly adapt pretext tasks originally developed for computer vision tasks, including multitask learning \cite{saeed2019multi} and masked reconstruction \cite{haresamudram2020masked}, where the model learns to predict masked parts of input data. Multitask learning pretext tasks rely on different signal processing-based data augmentations and transformations introduced by Um et al. \cite{um2017data} including rotations, flipping channels, time-warping, channel shuffling, and adding random Gaussian noise. These tasks have shown improvements in model performance by capturing principle signal characteristics without label information, thus reducing the dependency on labeled data. 

The motivation behind this paper stems from the observation that many of the pretext tasks devised for computer vision do not well reflect the characteristics of wearable sensor signals.  Information contained in sensor data is determined by  physical constraints such relationships between individual channels, gravity and biomechanical limitations, which are often ignored or even violated  in common computer vision transformations such  as rotation or scaling.  Thus, for example, given a 2D "pseudo image" representation of a multidimensional sensor data time window, an image rotation does not create a rotated sensor signal.  In fact, depending on the rotation parameters,  the result is likely not even to correspond to a valid sensor signal due to violation of physical dependencies between channels (e.g. between individual accelerometer channels and between accelerometer and gyro). This leads to suboptimal representations that may deteriorate model performance in HAR tasks, as they fail to capture the inherent physical dynamics of human movements.  

To address these limitations, in this paper, we propose a novel physics-informed multi-task pre-training (PIM) for SSL in wearable sensor-based HAR. The proposed PIM method is founded on three core aspects of human motion: the speed of movement, movement angles, and the symmetry between sensors placed at different body positions like on both wrists or limbs. By leveraging physics equations, we extract these features as pretext tasks that incorporate the fundamental physical characteristics of human activities. The integration of these physics-driven tasks within SSL enables the model to better capture realistic motion patterns and improves performance on HAR tasks. Experimental evaluations across four widely-used HAR benchmark datasets, including DSADS \cite{altun2010comparative}, PAMAP2 \cite{reiss2012introducing}, MM-Fit \cite{stromback2020mm}, and WEAR \cite{bock2023wear}, demonstrate that our approach achieves superior performance compared to state-of-the-art methods including multitask learning and masked reconstruction. 

The contributions of this work can be summarized as follows.
\begin{itemize}
    \item We introduce a novel framework for self-supervised learning in wearable sensor-based HAR that leverages physics-driven pretext tasks, specifically designed to account for the physical constraints of human motion.

    \item We identify and formulate three main aspects of human movement - speed, movement angles, and symmetry degrees between sensor locations - as key features for SSL, calculated using physics equations.

    \item Extensive experiments demonstrate the effectiveness of the proposed pretext tasks on four HAR benchmark datasets. The proposed method improves the performance of HAR and achieves the best performance among state-of-the-art methods, being specially effective when the number of examples per class is very limited, such as a few-shot learning scenario.

    \item We have conducted extensive ablation studies in order to verify the effectiveness of each type of pseudo-label, both alone as well as in combination. 
\end{itemize}


The remainder of this paper is organized as follows: \cref{sec:relatedwork} reviews related work, providing context for our approach. \cref{sec:method} presents a detailed introduction to the proposed physics-driven pretext tasks for SSL in HAR. In \cref{sec:experimentalresults}, we evaluate the proposed method across multiple HAR benchmark datasets, comparing its performance with state-of-the-art methods, and we discuss the limitations of the proposed method and future work. Lastly, \cref{sec:conclusion} concludes this paper.

\section{Related Work}
\label{sec:relatedwork}
Self-supervised learning (SSL) has emerged as a promising approach to address the challenges associated with labeled data scarcity in wearable sensor-based HAR. Collecting labeled data for HAR often requires controlled, lab-based studies where participants perform a set of pre-defined activities while wearing sensors, with video recordings used to synchronize and annotate the data streams accurately \cite{roggen2010collecting,ciliberto2021opportunity++}. This process, while effective, is costly, time-intensive, and raises privacy concerns \cite{fortes2021translating}. As a result, publicly available datasets are typically limited in scope, involving only a small number of participants and activity types, which constrains the development of more complex deep learning models for complicated activity recognition. 

Unlike labeled data, large quantities of unlabeled sensor data can be collected relatively easily in real-world settings, where wearable devices like smartwatches can passively record data from numerous participants over extended periods. Studies like the UK Biobank for wearable sensor-based HAR have demonstrated the feasibility of large-scale, unlabeled data collection, generating significant volumes of real-world sensor data from tens of thousands of participants \cite{doherty2017large,willetts2018statistical,yuan2024self}. Such approaches not only increase the scale of data but also introduce diverse activity patterns and participant demographics. However, changes in sensor technology and distribution shifts over time further emphasize the need for methods that can learn robust representations without requiring annotated data.

SSL enables HAR models to leverage this abundance of unlabeled data by learning useful representations through pretext tasks. In the pretrain-then-finetune framework, the network is first pre-trained on unlabeled data using pretext tasks designed to capture essential characteristics of the data \cite{fortes2023don,nshimyimana2025contrastive}. Once pre-trained, the learned weights can be fine-tuned on a smaller labeled dataset, adapting them for specific downstream tasks, such as activity recognition.

One popular SSL approach in HAR is multi-task learning with data augmentations, where transformations like Gaussian noise addition, scaling, rotation, and time-warping are applied to the data \cite{um2017data,saeed2019multi}. The network learns to classify whether each transformation was applied, capturing core signal characteristics across varying conditions \cite{saeed2019multi}. Another powerful SSL approach is masked reconstruction. Masked reconstruction \cite{haresamudram2020masked} withholds portions of the data and trains the transformer network \cite{vaswani2017attention} to reconstruct the missing parts, encouraging the learning of contextual representations based on available data segments. This helps networks learn patterns across channels and time by simulating a denoising effect. 

SSL for wearable sensor-based HAR continues to be a promising research area, as these pretext tasks increasingly incorporate the unique characteristics of sensor data \cite{haresamudram2022assessing,jain2022collossl}. However, existing methods often adapt techniques from computer vision, without addressing the specific physical constraints of HAR data. Consequently, our work builds on SSL for HAR by proposing a novel approach that considers these physical factors, developing pretext tasks aligned with the physics underlying human motion.

\section{Methodology}
\label{sec:method}
In this section, we describe our self-supervised learning framework for wearable-based HAR, which uses physics-driven physical quantities related to human motions for effective representation learning through pre-training. More specifically, we use the speed of movement, rotation, and degree of synchrony between signals to create pseudo-labels for unlabeled data. Our main idea is that learning to predict those can lead to a good representation of HAR, since those characteristics are relevant for a broad range of activities. 

\begin{figure*}
    \centering
    \includegraphics[width=1\textwidth]{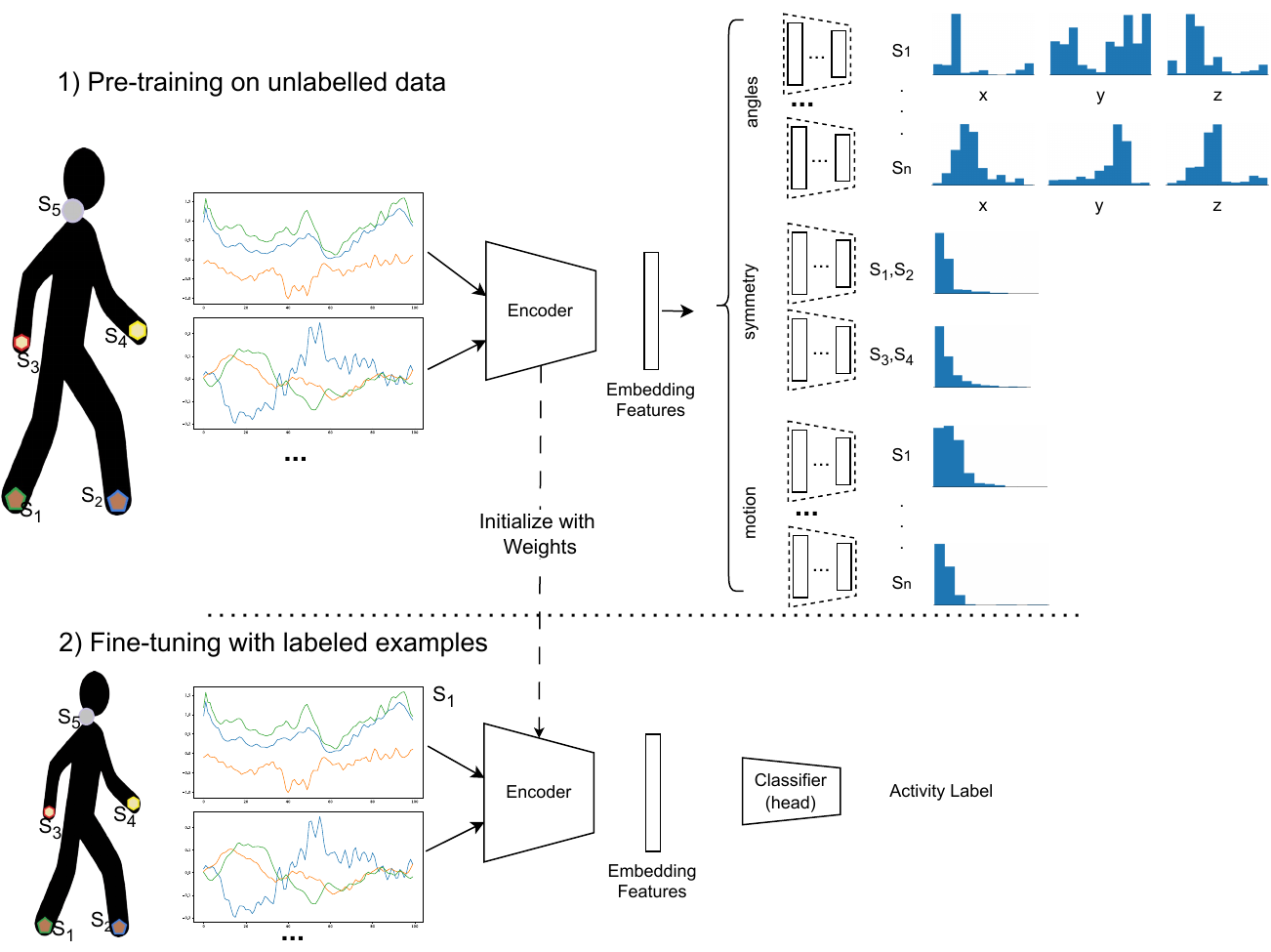}
    \caption{Our Physics-Informed Pre-training framework for learning better representations for HAR.  We have developed relevant physical quantities related to human motions that can be computed from unlabeled sensor data. Our pre-training consists of learning to predict discretized versions of those quantities. This encoder can then be fine-tuned together with a classification head using a small amount of labeled data to train a HAR classifier.}
    \label{fig:method}
    \vspace{-5mm}
\end{figure*}

Our framework consists of two steps, as depicted in \cref{fig:method}. First, unlabeled data is used to pre-train an encoder by predicting the pseudo-labels. Each pseudo-label is predicted by its own dedicated head network, and its classes represent ranges in our proposed physically-relevant quantities.
We also experimented with directly predicting the values of our developed features through regression, but obtained better results by discretizing each one and treating them as pseudo-classes. After pre-training on unlabeled data, we add a HAR classification head to the encoder and fine-tune both with a small number of unlabeled examples of the desired classes. Finally, our network is tested on unseen data for evaluation.

More formally,  Let $X_u = [x_1, ..., x_n]$ be the time-series sensor data obtained by applying a sliding window of size $n_w$ of the $n_c$ available sensor channels with $x_i \in \mathbb{R}^{n_c \times n_w} $. This data is not labeled, and thus our method first computes $k$ \textbf{pseudo-labels} per window, meaning $p_i \in \mathbb{R}^{k}$, creating our set of pseudo-labels $P_Y = [p_1, ..., p_n]$. Using $X_u$ and $P_Y$ we perform our pretext task. As depicted in \cref{fig:method} sensor data passes first through an $enc$ network $\mathbb{R}^{n_c \times n_w}  \rightarrow \mathbb{R}^r $ where $r$ is the size of our embedding. This representation is then used to predict the pseudo-labels using multiple head networks, with head $H_j$ performing $H_j(enc(x))$ to predict pseudo-label $j$. Once we are done with the pretext training, we attach a classification head $C$ to our pre-trained $enc$, so that $C(enc(x))$ predicts the activity labels present in the target HAR dataset. This network is then trained using a small set of labeled examples $X_l = [x_{l_1}, ..., x_{l_n}]$ with $x_{l_i} \in \mathbb{R}^{n_c \times n_w} $ that now have labels $Y = [y_1, ..., y_n]$ the corresponding activity label set of $X_l$. After this fine-tuning is performed, $C(enc(x))$ is evaluated on a set of unseen users of the target dataset, whose data is not part of either $X_u$ or $X_l$.

\subsection{Pretext Task Definition}
At the heart of our method are our proposed physic-driven quantities, each encoding a fundamental aspect of human motion. We separate our features between those related to motion velocities, rotations, and symmetry. In this Section, we will describe our pseudo-labels, their motivation, and how they are computed.


\subsubsection{Speed of Motion}
Movement speed is a fundamental aspect of human motion. This is clear not only for locomotion classes (such as sitting still versus walking), but also for many other domains. This is highlighted by the ubiquity of acceleration as a modality for HAR. Still, acceleration is affected by sensor orientation and placement. In order to better learn representations related to the speed of motion we developed pseudo-labels related to the linear speed of motion. We pre-process acceleration data to obtain the norm of the linear speed, as this is more representative of limb motion regardless of gravity and sensor orientation.

\begin{figure*}
    \centering
    \includegraphics[width=1.0\linewidth]{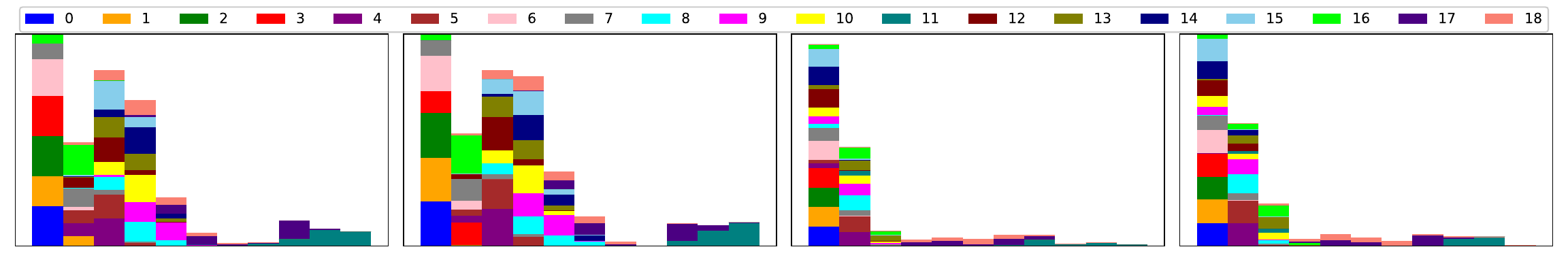}
    \caption{Histogram of the speed of motion pseudo-labels computed on DSADS dataset. We have colored the histograms per class to show the different distribution of speed features per class. As we can see, there some classes (such as 11) have high speeds, while many others have speed values in other ranges.}
    \label{fig:ps:speed}
\end{figure*}

The full procedure is as follows. For each sensor position, unlabeled acceleration sliding windows are used to compute:
\begin{enumerate}
    \item An approximation of gravity removal is applied by passing the signal through a Butterworth filter with 2Hz cutoff.
    \item From the obtained linear acceleration $a_{x,y,z}$ we compute speeds for each timestep $k$ using 
    \begin{equation}
        v_{x,y,z}[k+1] = v_{x,y,z}[k] + a_{x,y,z}[k+1] * \Delta s
    \end{equation}
    from data sampled at frequency $1 \mathbin{/} \Delta s$.
    \item Then we obtain positions using
    \begin{equation}
         t_{x,y,z}[k+1] = t_{x,y,z}[k] + v_{x,y,z}[k] * \Delta s      \\
    \end{equation}
    \item This sequence of positions is further de-noised by applying a Butter filter.
    \item Finally, we compute our orientation-independent speed of motion feature using the Euclidean norm of the speed:
        \begin{equation}
            \Delta D_{x,y,z}^w = \sqrt{\sum \left( t_{x,y,z}[i+1] - t_{x,y,z}[i] \right)^2} 
        \end{equation}
\end{enumerate}
As we can see in \cref{fig:ps:speed}, the speed of motion is highly informative for many movement classes. Once we have computed this feature for each sensor position, each one is independently converted into a pseudo-label by discretizing it into $11$ uniform intervals. This means we include one pseudo-label classification pretext task per sensor position: predicting its interval of speed motion. Each one of these tasks is done by an independent network head, as can be seen in \cref{fig:method}.

\subsubsection{Angular features}
The angle of motions for the limbs is relevant in many motions. For example, the supination and pronation of the forearm are involved in the motion of turning a doorknob. This is relevant to fitness exercises, manual work (consider turning a screw), and beyond. Unfortunately, the full set of sensing modalities for estimating reliably the full orientation of a sensor. In fact, it is common for devices to include only accelerometers, lacking gyroscopes, and magnetometers in their IMUs. Having this in mind, our angle features were developed to also encompass cases where only acceleration is available, but taking advantage of other modalities when available.

In case there is only accelerometer data for a sensor, we will compute its pseudo-labels based on the gravity angle alone, similar to \cite{plotz2011feature,karantonis2006implementation}. The full procedure is as follows:
\begin{enumerate}
    \item Estimate the gravity vector by applying a Butterworth filter with 2Hz cutoff.
    \item Angles computed for each timestamp $k$ using our estimated gravity vector $g$ using:
    \begin{equation}
        \begin{split}
            \phi[k] = arctan2(g_y[k],g_z[k]) \\
            \theta[k] = arctan2(g_x[k],g_z[k]) \\ 
            \psi[k] = arctan2(g_y[k],g_x[k]) \\
        \end{split}
    \end{equation}
    where $\phi,\theta,\psi$ represent the roll, pitch and yaw, respectively.
    
    \item This provides angles in the $(-\pi, \pi)$ range. 
    Since we are interested in converting angles sequences to single values that represent intervals in the sense of 'low' or 'high' angle present for a time window, it is not desirable for negative and positive angles to cancel each other when computing the mean angle in the window. Thus, we will use absolute angle values. In order to also include the overall direction of movement direction, we will also multiply values by its dominant sign in the window. Thus, features are computed using
    \begin{equation}
        \begin{split}
            \Delta R_{x} &= \frac{1}{N} \sum^{N}_k |\phi[k]| \times sign(\sum^{N}_k \phi[k])  \\
            \Delta R_{y} &= \frac{1}{N} \sum^{N}_k |\theta[k]| \times sign(\sum^{N}_k \theta[k])  \\
            \Delta R_{z} &= \frac{1}{N} \sum^{N}_k |\psi[k]| \times sign(\sum^{N}_k \psi[k]) \\
        \end{split}
    \end{equation}
    where 
    \[
    sign(x) = 
    \begin{cases}
        1 & \text{for } x \geq 0 \\
        -1 & \text{for } x < 0 \\
    \end{cases}
  \]
\end{enumerate}
\begin{figure*}
    \centering
    \includegraphics[width=1\linewidth]{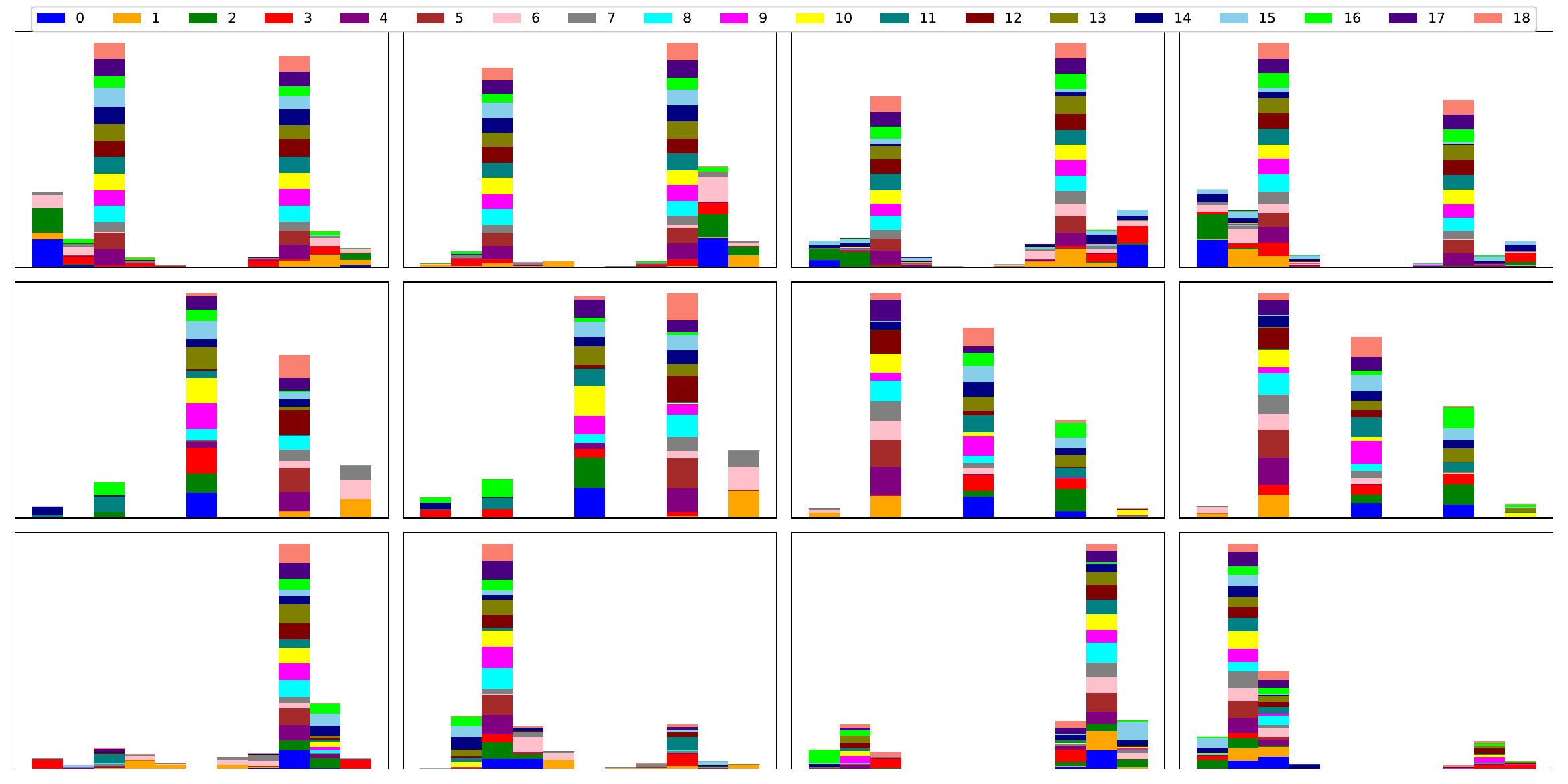}
    \caption {Distribution of angle physical quantities in the DSADS dataset for four devices. Rows represent roll, pitch and yaw while columns represents the different sensor positions. Histograms are colored by class to show pseudo-label distribution.} 
    \label{fig:ps:angle:dsads}
\end{figure*}

In case we have access to more modalities, such as gyroscope and magnetometer, then we use those for a better estimation of the angles at each timestep by applying the approach proposed in \cite{Ellis2016:features:angle} and MadgwickAHRS \cite{githubGitHubMorgilmadgwick_py}. After that, the same feature computation is performed.

Finally, we defined ten thresholds, evenly dividing the range $[-\pi..\pi]$ into eleven equal intervals, each with fixed size $0.628$, as we already know the range of this feature. Each sensor placement thus generates three distinct pseudo-labels. In the pretraining each pseudo-label feature is forwarded to its independent head network. \cref{fig:ps:angle:dsads} shows the information computed from the angles.

\subsubsection{Symmetry (synchronous) behavior}
\begin{figure}
    \centering
    \includegraphics[width=1\linewidth]{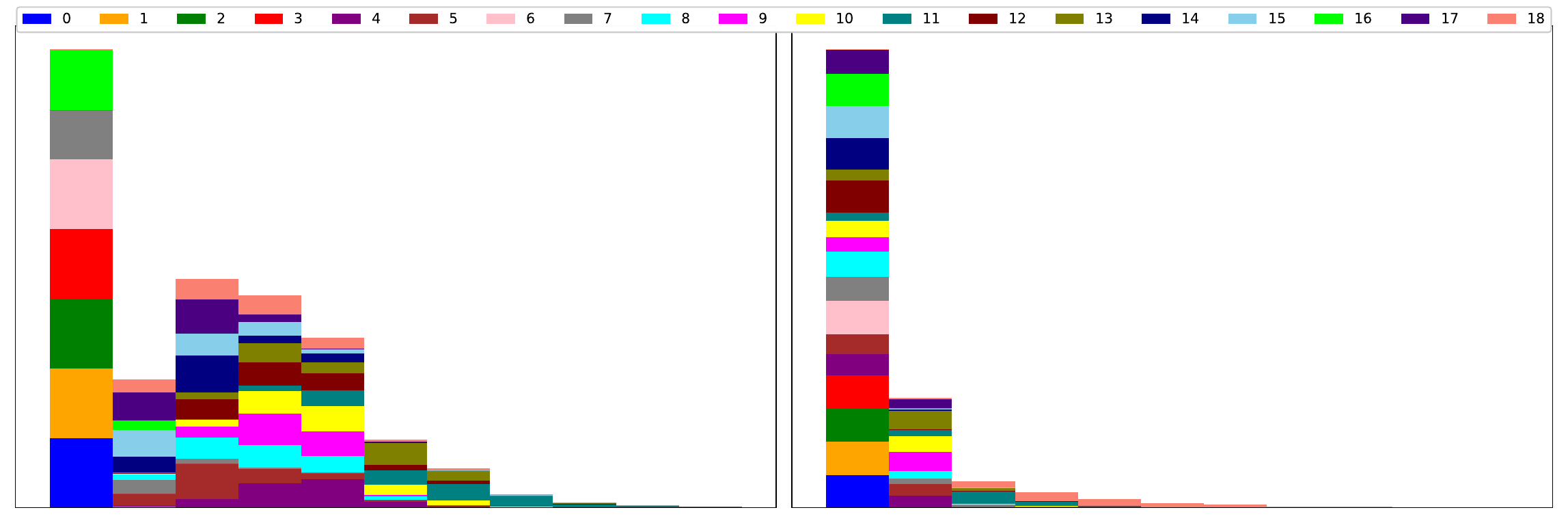}
    \caption{ Histogram of pseudo-labels related to synchronization in the DSADS dataset for the legs (on the left) and arms (on the right). The histograms were colored per class to show how different classes fall into different bins for our pseudo-labels. }
    \label{fig:ps:synch:dsads}
\end{figure}

Symmetry (or lack thereof) is also a relevant aspect of human motions. Synchronized motions include walking, running, or swimming while examples of unsynchronized ones encompass writing, throwing a ball, or waving. In our framework, we measure the symmetry as the distance between acceleration signals  $a_{x_1}$ and $a_{x_2}$ from two devices. We include two cases: left-right arms symmetry and left-right legs symmetry. Their effect is illustrated in \cref{fig:ps:synch:dsads}. The computation of $\Delta D_{symmetry}$ is done on those pairs as follows:

\begin{enumerate}
    \item First we filter $a_{x_1}$ and $a_{x_2}$ by applying a Butterworth filter with $2$ Hz cutoff.

    \item Then we compute the root squares of those two signals, making each of them a one-dimensional time series:
    \begin{equation}
        x_1[k] = \sqrt{\sum \limits_{c}^{} a_{x_1,c}[k]^2};  \qquad
        x_2[k] = \sqrt{\sum \limits_{c}^{} a_{x_2,c}[k]^2}  \\ 
    \end{equation}

    \item Then we align signals by finding the segment with maximum cross-correlation between the two sequences. 
    \begin{equation}
        \begin{split}
            shift &= argmax(correlate(x_1, x_2)) \\
            x_{1} &= align(x_1, shift) \\
            x_{2} &= align(x_2, shift) \\
        \end{split}
    \end{equation}


    \item Finally, the dynamic time warping (DTW) distance \cite{kale2012impact:dtw:trajectory} is applied to what remains of the signals, making our symmetry feature as shown in \cref{fig:ps:synch:dsads}.
    \begin{equation}
       \Delta D_{symmetry} = DTW(x_1, x_2)
    \end{equation}
\end{enumerate}

\begin{figure*}[!t]
    \label{fig:tsne}
    \begin{subfigure}[b]{0.48\textwidth}
         \centering
         \includegraphics[width=210pt]{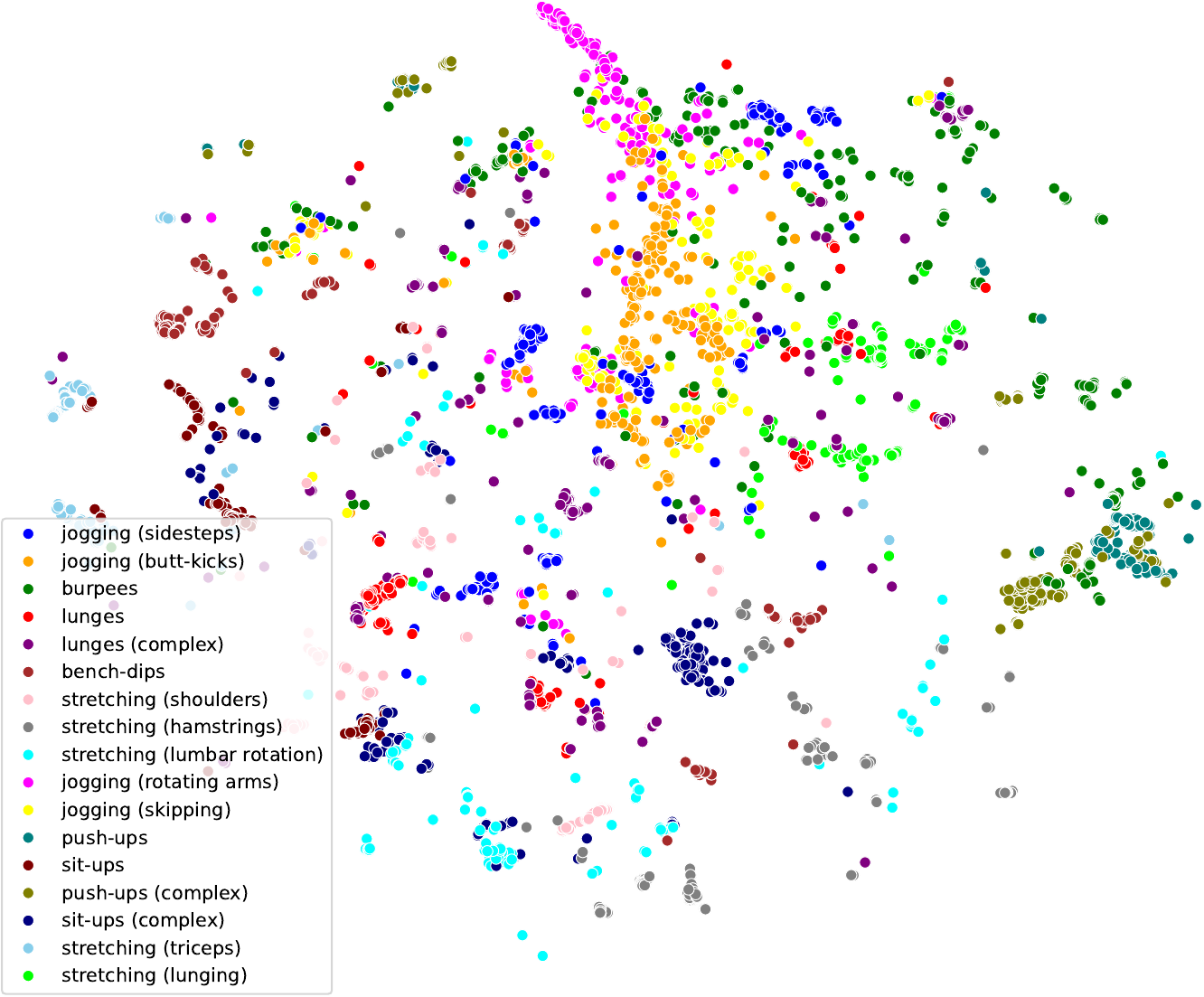}
         \caption{WEAR dataset (removed one dominating class for clarity).}
         \label{fig:tnse:wear}
     \end{subfigure}
     \hfill
     \begin{subfigure}[b]{0.48\textwidth}
         \centering
         \includegraphics[width=210pt]{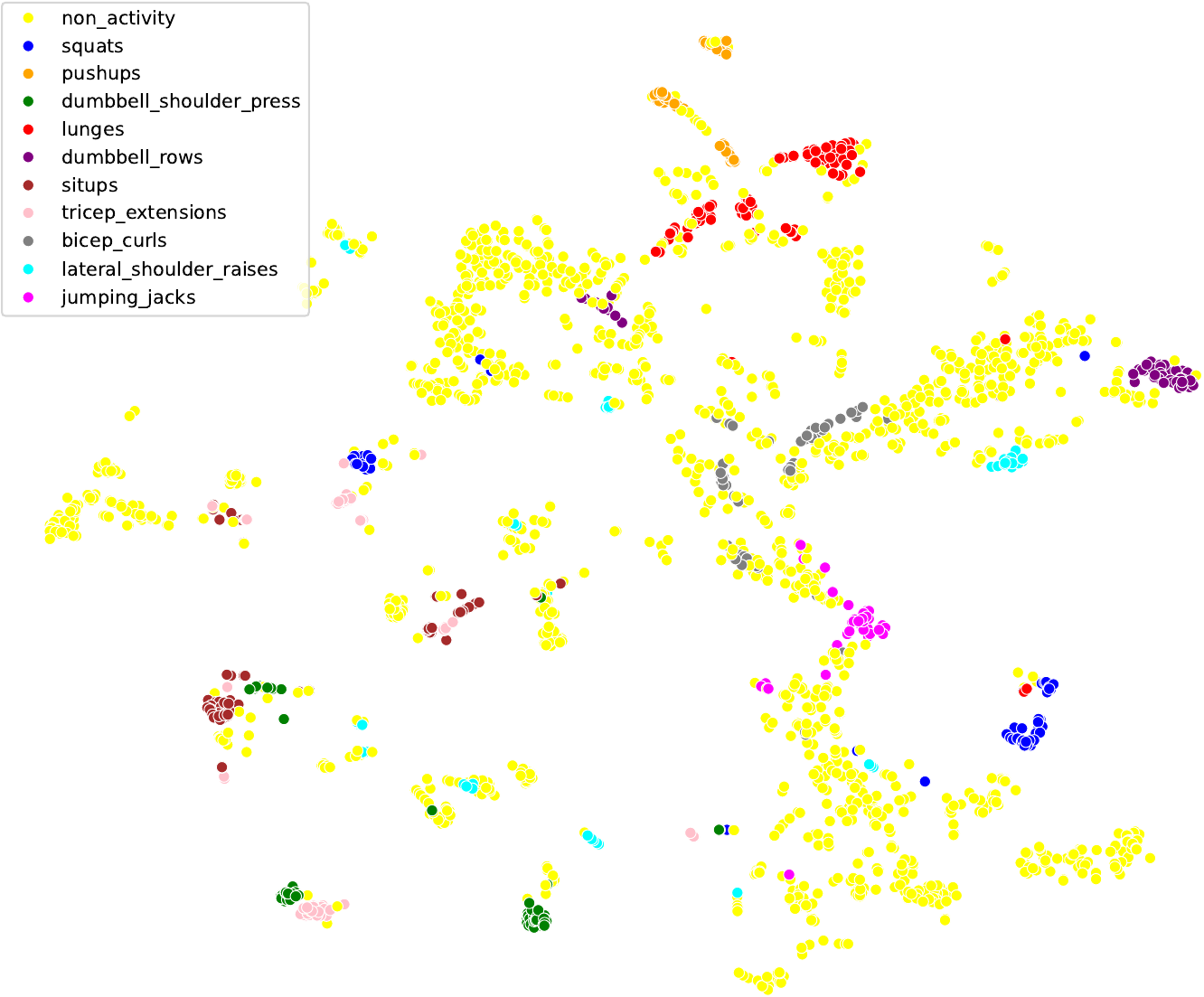}
         \caption{MM-Fit dataset.}
         \label{fig:tsne:mmfit}
    \end{subfigure}
        \\
    \begin{subfigure}[b]{0.48\textwidth}
         \centering
         \includegraphics[width=210pt]{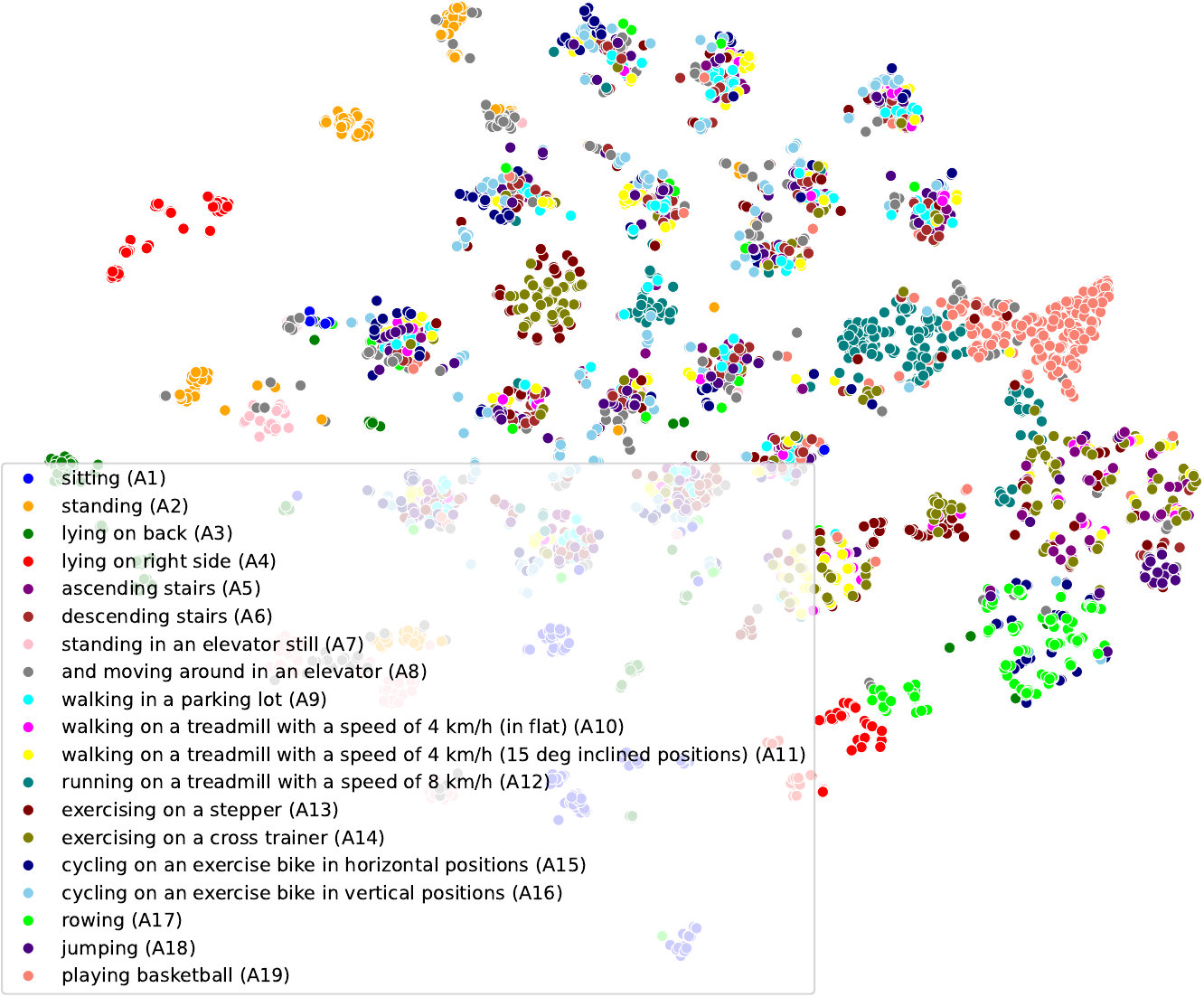}
         \caption{DSADS dataset.}
         \label{fig:tsne:dsads}
     \end{subfigure}
     \hfill
     \begin{subfigure}[b]{0.48\textwidth}
         \centering
         \includegraphics[width=210pt]{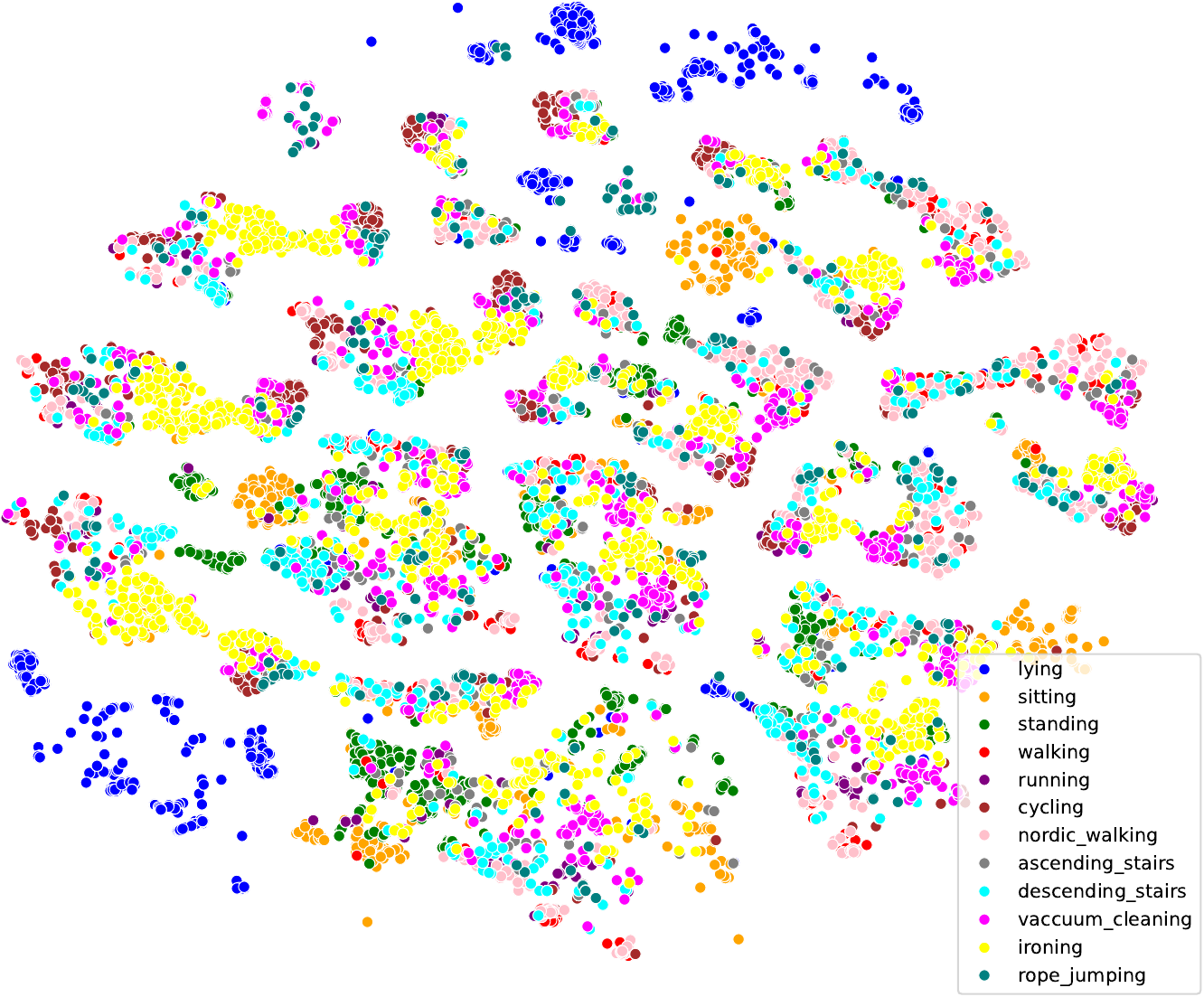}
         \caption{PAMAP2 dataset.}
         \label{fig:tsne:pamap2}
     \end{subfigure}
     \caption{T-SNE of the physics-driven quantities before clustering, with points colored by activity class.}
     \vspace{-5mm}
\end{figure*}

As we can see in \cref{fig:tsne}, where we visualize our physics-inspired quantities for each point in the studied datasets, those pseudo-labels do not fully distinguish between all HAR classes. What they create is a useful space where some classes are well separated (such as some types of jogging in the WEAR dataset or rowing in DSADS). Thus, learning our pseudo-labels prepares the network by starting from an informative space.

\subsection{Self-Supervised Learning with pseudo-class}
The proposed method learns from three properties in IMU-data termed SAM-task
\begin{inparaenum}[i)]
\item \textbf{S}ymmetry of the human body
\item \textbf{A}ngular features of moving device
\item Speed of \textbf{M}otion.
\end{inparaenum}
To enable the use of binary and categorical cross-entropy in self-supervised learning, the features are discretized into eleven uniform bins using the KBinsDiscretizer from scikit-learn \cite{pedregosa2011scikit}. 

Each one of the tasks has its own loss, meaning we have $\mathcal{L}_{angle}$, $\mathcal{L}_{motion}$ and $L_{symmetry}$, respectively. 
For the angular features loss $\mathcal{L}_{angle}$, SAM-task is defined through multi-label classification using binary cross entropy, whereas the other two losses use categorical cross-entropy. 

In summary, pre-training was guided by aggregated loss function as in the equation Eq. \ref{eq:loss} where the default value for $\alpha$, $\beta$ and $\gamma$ were set to one in our experiments. Alternatively, these parameters can be set as learnable. Pseudo-label provides a different number of classes depending on the dataset configuration to train twelve SAM-task heads for SSL.  For a dataset with $n$ sensors there are  $3*n*11$ classes for angle features, $n*11$ for speed of motion and $d*11$ for symmetry where $d$ is the the number of pairs of limbs where synchrony is computed (so $d$ is 1 if only arms (or legs) and 2 if arms and legs).
\begin{equation}
    \label{eq:loss}
    \mathcal{L} = \alpha \times \mathcal{L}_{symmetry} + \beta \times \mathcal{L}_{angle} + \gamma \times \mathcal{L}_{motion}
\end{equation}

\section{Experimental Results}
\label{sec:experimentalresults}

To ensure a fair comparison on accuracy, the proposed and other state-of-the-art methods were evaluated under the same settings, including architecture, dataset splits, and hyperparameters.  Fine-tuning was performed by initializing from pretrained weights without freezing based on \cite{fortes2023don}. Furthermore, to evaluate the performance of pretraining methods, we varied the amount of labeled data available during fine-tuning. For initialization, we created different ten random weight initializations for the shared block, so all training started from consistent initial conditions. The pseudo-label-based method was compared with the baseline, multi-task \cite{saeed2019multi} and masked reconstruction \cite{haresamudram2020masked}. 

\subsection{Datasets and Evaluation Metrics}
To evaluate the effectiveness of the proposed method, we evaluate it on four representative benchmark HAR datasets: MM-Fit \cite{stromback2020mm}, WEAR \cite{bock2023wear}, PAMAP2 \cite{reiss2012introducing}, and Daily and Sports Activities Dataset (DSADS) \cite{altun2010comparative}.

\begin{itemize}
    \item \textbf{MM-Fit dataset} \cite{stromback2020mm} is an open-source dataset collected from various body-worn inertial sensors and external cameras, all time-synchronized. It includes data from four types of devices: an Orbbec Astra Pro camera, an eSense earbud, two Mobvoi TicWatch Pro smartwatches, and two smartphones (Samsung S7 and Huawei P20). The participant wore smartwatches on both wrists, an earbud in the left ear, and smartphones in both left and right trouser pockets. The smartwatches recorded acceleration and angular velocity at 100 Hz, and for this study, the accelerometers from the smartwatches were used. The dataset contains 10 workout activities such as squats, lunges, bicep curls, sit-ups, push-ups, and others, along with a 'no activity' class. 
    In our experiment, we focused exclusively on the workout data where all IMU sensor streams were available. Since we could only find the workout session, any partition on this dataset is based on workout IDs. For pretraining, we trained all SSL methods with workout sessions $3$, $14$, and $14$, and evaluated using $8$ workout sessions. The remaining data was used for fine-tuning and for the fully supervised baseline. 

    \item \textbf{WEAR dataset} \cite{bock2023wear} is a multimodal dataset for outdoor sports activity recognition, featuring both inertial data (accelerometers) and egocentric video data. The dataset includes recordings from 18 participants who performed 18 different workout activities, covering a variety of sports-related activities such as different styles of running, stretching, and strength-based exercises. These activities were recorded across 10 distinct outdoor locations, ensuring diverse environmental contexts. The wearable sensors include accelerometers placed on both limbs and a camera mounted on the head to capture video. The inertial data was sampled at 50Hz, providing high-frequency motion capture for precise activity recognition. The data was partitioned by subject for the entire experiment, unless otherwise specified. For SSL training, we used subjects $10, 11, 12, 18, 19, 20, 21$, along with two additional unlabeled sessions. The remaining subjects were used for fine-tuning and fully supervised baseline classification.

    \item \textbf{PAMAP2 dataset} \cite{reiss2012introducing} contains sensor recordings from nine participants performing 18 daily activities, totaling over 10 hours of data. For data collection, participants wore a heart rate monitor and three inertial measurement units (IMUs) placed on the chest, dominant wrist, and dominant ankle. Each IMU captures 17 channels of data, resulting in 52 channels in total, including a single heart rate channel. The IMU data consist of 6 channels of acceleration, 3 channels of gyroscope, 3 channels of magnetometer, and 3 channels of orientation. We selected data from eight participants, retaining 12 activities categorized under the "protocol" subset, while excluding the "optional" activities recorded for only one participant and omitting the ninth participant who performed only one activity. To preprocess, all NaN values were filled using linear interpolation, and sensor data was normalized to zero mean and unit variance using mean and standard deviation computed from training subset. We used a sliding window of 2 seconds (200 samples) with a 0.5-second step (50 samples). The pretraining was performed on $2, 4, 6, 8$ subjects and classification utilized $1, 3, 5, 7$. The leave-one-subject-out evaluation scheme was adopted for this dataset with downstream subjects. 

    \item \textbf{DSADS dataset} \cite{altun2010comparative} comprises sensor data from eight participants (4 female and 4 male, ages 20 to 30) performing 19 distinct activities. Each activity was performed for a duration of 5 minutes, resulting in approximately 480 segments per activity (60 segments per participant) by dividing the recordings into 5-second windows. Participants were encouraged to perform each activity according to their own style without constraints on speed or intensity, allowing for inter-subject variations in activity execution.
    We utilized $2, 4, 6, 8$ subjects for SSL and $1, 3, 5, 7$ subjects for classification.
    
\end{itemize}

\textbf{Metrics.}   
Two common metrics are used to study the performance of the models: macro F1 and accuracy. We run every experiment ten times and report the mean and standard deviation. Before any training, the dataset was partitioned based on subjects into two subsets for pretraining and classification, respectively. The subset for pretraining was again split into train-validation sets, using a random 70-30 splits of the segmented data points. For fine-tuning (and fully supervised baseline) we applied the leave one subject out (LOSO) cross-valuation scheme on subjects from classification subset.

\subsection{Implementation Details}

\begin{figure*}
    \centering
    \includegraphics[width=1\linewidth]{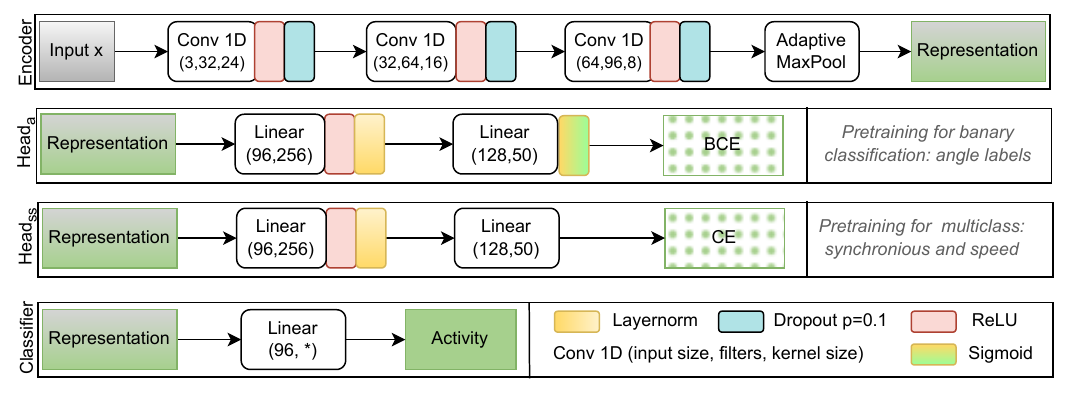}
    \caption{Show the network architecture. From top the encoder which is common for all the experiment. $Head_b$ is the classification head for binary cross entropy used for angle features while the $Head_{ss}$ predicts the synchronous and speed pseudo classes.}
    \label{fig:nets-architecture}
\end{figure*}

\paragraph{Encoder}
In all cases (and for all methods) our encoder is a CNN which consists of three 1D convolutional layers with 32, 64, and 96 feature maps, using kernel sizes of 24, 16, and 8, respectively, with a stride of 1. A dropout rate of 0.1 is applied after each layer to prevent overfitting. Global max pooling is employed after the final convolutional layer to aggregate high-level discriminative features. Further details can be found in the Fig \ref{fig:nets-architecture}.

\paragraph{Pretraining}
The pre-training implementation considers a shared encoder (common layers) and distinct heads (task-specific layers) for each SAM-task. Each SAM-task is associated with its own loss function, and the shared block encourages efficient weight utilization across tasks by capturing generalized representations through the shared layers \cite{saeed2019multi}.

Each task-specific branch includes two fully connected layers with 256 and 128 hidden units, respectively, separated by layer normalization and a sigmoid output layer for binary classification. Only the classification for symmetry and speed does not need a sigmoid. ReLU activation is used in all layers except for the output. The network is trained using the Adam optimizer for a maximum of 100 epochs, with a learning rate of 0.0004 unless specified otherwise.
 
Our training process is summarized in \cref{fig:method}. For each input instance, we first generate pseudo-labels of each data point (window signal) for the self-supervised pre-training of the network. Our method is compatible with data augmentations such as horizontal flipping, axis permutation, and time warping, as implemented in \cite{tang2020exploring}. Thus, we increase the amount of training data by generating three augmented versions of each training sample (but without changing its pseudo-label). The augmentations selected from \cite{tang2020exploring} were permutation, time-warping and horizontal flipping. We oversample the original points so that there are as many non-augmented as augmented ones in the training set. During training, model weights with the best validation loss from the pretraining phase are saved to ensure optimal performance in downstream-task.

\paragraph{Downstream task: Classification and Finetuning}
For activity recognition, the same network architecture is used for all methods to demonstrate the performance of each training strategy independently of the architecture. The encoder (as defined above) is utilized to extract features, which are then taken as input by the classifier head to predict the classes. The classifier head consists of a single linear layer (see Fig. \ref{fig:nets-architecture} for details), as more complex architectures did not show any improvement.



\begin{table*}[!t]
\caption{comparison of state-of-the-art pretraining methods with the proposed physics-informed multi-task pre-training (PIM) approach by changing the number of available labeled data for fine-tuning from 2 to 8 examples per class.}
\centering
\resizebox{\textwidth}{!}{%
\begin{tabular}{ll|ll|ll|ll}
\multirow{2}{*}{Dataset} & \multirow{2}{*}{\begin{tabular}[c]{@{}l@{}}\# of Available labels/\\ Pretrain method\end{tabular}} & \multicolumn{2}{c|}{2}                      & \multicolumn{2}{c|}{4}                      & \multicolumn{2}{c}{8}                                            \\
                         &                                                                                                    & F1                   & ACC                  & F1                   & ACC                  & F1                   & ACC                 \\ \hline
\multirow{4}{*}{PAMAP2}  & Baseline                                                                                           & 0.264±0.153          & 0.33±0.155           & 0.3±0.193            & 0.365±0.198          & 0.44±0.187           & 0.511±0.17           \\
                         & Masked Reconstruction                                                                              & 0.277±0.166          & 0.342±0.165          & 0.283±0.165          & 0.346±0.169          & 0.445±0.195          & 0.511±0.175          \\
                         & Multitask                                                                                          & 0.276±0.129          & 0.338±0.128          & 0.377±0.155          & 0.453±0.148          & 0.447±0.177          & 0.516±0.156          \\
                         & PIM                                                                                  & \textbf{0.351±0.141} & \textbf{0.409±0.133} & \textbf{0.395±0.176} & \textbf{0.463±0.17}  & \textbf{0.541±0.176} & \textbf{0.613±0.131} \\ \hline
\multirow{4}{*}{WEAR}    & Baseline                                                                                           & 0.19±0.085           & 0.199±0.085          & 0.258±0.104          & 0.25±0.097           & 0.387±0.128          & 0.366±0.124          \\
                         & Masked Reconstruction                                                                              & 0.179±0.098          & 0.192±0.100            & 0.254±0.104          & 0.251±0.098          & 0.383±0.121          & 0.359±0.122          \\
                         & Multitask                                                                                          & 0.187±0.094          & 0.192±0.094          & 0.274±0.100            & 0.263±0.097          & 0.382±0.130           & 0.358±0.121          \\
                         & PIM                                                                                  & \textbf{0.214±0.079} & \textbf{0.218±0.082} & \textbf{0.32±0.082}  & \textbf{0.307±0.084} & \textbf{0.452±0.109} & \textbf{0.423±0.108} \\ \hline
\multirow{4}{*}{MM-Fit}   & Baseline                                                                                           & 0.172±0.107          & 0.198±0.11           & 0.278±0.111          & 0.237±0.106          & 0.405±0.123          & 0.347±0.118          \\
                         & Masked Reconstruction                                                                              & 0.172±0.11           & 0.201±0.118          & 0.218±0.141          & 0.244±0.141          & 0.423±0.14           & 0.345±0.131          \\
                         & Multitask                                                                                          & 0.172±0.087          & 0.215±0.103          & 0.229±0.114          & 0.263±0.131          & 0.399±0.096          & 0.363±0.093          \\
                         & PIM                                                                                  & \textbf{0.269±0.086} & \textbf{0.254±0.073} & \textbf{0.328±0.088} & \textbf{0.271±0.078} & \textbf{0.441±0.063} & \textbf{0.417±0.076} \\ \hline
\multirow{4}{*}{DSADS}   & Baseline                                                                                           & 0.317±0.106          & 0.376±0.107          & 0.394±0.144          & 0.455±0.137          & 0.535±0.073          & 0.585±0.062          \\
                         & Masked Reconstruction                                                                              & 0.335±0.133          & 0.391±0.129          & 0.393±0.117          & 0.45±0.114           & 0.538±0.088          & 0.585±0.079          \\
                         & Multitask                                                                                          & 0.348±0.159          & 0.402±0.16           & 0.447±0.095          & 0.5±0.09             & 0.544±0.095          & 0.589±0.086          \\
                         & PIM                                                                                  & \textbf{0.445±0.101} & \textbf{0.504±0.098} & \textbf{0.518±0.09}  & \textbf{0.574±0.08}  & \textbf{0.592±0.075} & \textbf{0.636±0.064}
\end{tabular}
}
\label{tab:comparison_results1}
\end{table*}

\begin{table*}[!t]
\caption{comparison of state-of-the-art pretraining methods with the proposed physics-inspired multi-task pre-training (PIM) approach by changing the number of available labeled data for fine-tuning from 16 per HAR class to all per class.}
\centering
\resizebox{\textwidth}{!}{%
\begin{tabular}{ll|ll|ll|ll|ll}
\multirow{2}{*}{Dataset} & \multirow{2}{*}{\begin{tabular}[c]{@{}l@{}}\# of Available labels/\\ Pretrain method\end{tabular}} & \multicolumn{2}{c|}{16}                     & \multicolumn{2}{c|}{32}                     & \multicolumn{2}{c|}{64}                     & \multicolumn{2}{c}{All}                     \\
                         &                                                                                                    & F1                   & ACC                  & F1                   & ACC                  & F1                   & ACC                  & F1                   & ACC                  \\ \hline
\multirow{4}{*}{PAMAP2}  & Baseline                                                                                           & 0.539±0.181          & 0.612±0.146          & 0.574±0.176          & 0.647±0.135          & 0.582±0.184          & 0.654±0.143          & 0.593±0.186          & 0.663±0.145          \\
                         & Masked Reconstruction                                                                              & 0.515±0.189          & 0.582±0.16           & 0.572±0.176          & 0.646±0.124          & 0.585±0.183          & 0.657±0.138          & 0.593±0.196          & 0.661±0.155          \\
                         & Multitask                                                                                          & 0.534±0.171          & 0.609±0.13           & 0.56±0.176           & 0.636±0.139          & 0.578±0.191          & 0.651±0.155          & 0.594±0.198          & 0.657±0.16           \\
                         & PIM                                                                                  & \textbf{0.567±0.179} & \textbf{0.634±0.14}  & \textbf{0.594±0.188} & \textbf{0.669±0.138} & \textbf{0.6±0.187}   & \textbf{0.677±0.138} & \textbf{0.609±0.197} & \textbf{0.68±0.156}  \\ \hline
\multirow{4}{*}{WEAR}    & Baseline                                                                                           & 0.473±0.137          & 0.439±0.13           & 0.529±0.152          & 0.499±0.14           & 0.568±0.16           & 0.541±0.146          & 0.681±0.192          & 0.728±0.156          \\
                         & Masked Reconstruction                                                                              & 0.467±0.143          & 0.432±0.134          & 0.53±0.15            & 0.495±0.137          & 0.566±0.163          & 0.537±0.145          & 0.677±0.192          & 0.726±0.155          \\
                         & Multitask                                                                                          & 0.469±0.148          & 0.44±0.134           & 0.53±0.156           & 0.502±0.144          & 0.577±0.166          & 0.549±0.148          & 0.686±0.195          & 0.736±0.157          \\
                         & PIM                                                                                  & \textbf{0.533±0.121} & \textbf{0.497±0.112} & \textbf{0.582±0.136} & \textbf{0.547±0.125} & \textbf{0.616±0.146} & \textbf{0.583±0.131} & \textbf{0.702±0.182} & \textbf{0.748±0.15}  \\ \hline
\multirow{4}{*}{MM-Fit}   & Baseline                                                                                           & 0.567±0.094          & 0.533±0.104          & 0.672±0.057          & 0.687±0.067          & 0.73±0.058           & 0.77±0.064           & 0.846±0.061          & 0.917±0.028          \\
                         & Masked Reconstruction                                                                              & 0.57±0.059           & 0.513±0.092          & 0.659±0.064          & 0.659±0.082          & \textbf{0.738±0.064} & \textbf{0.775±0.066} & 0.823±0.153          & 0.911±0.052          \\
                         & Multitask                                                                                          & 0.536±0.099          & 0.511±0.115          & 0.645±0.077          & 0.661±0.098          & 0.731±0.072          & 0.775±0.074          & \textbf{0.853±0.054} & \textbf{0.922±0.024} \\
                         & PIM                                                                                  & \textbf{0.571±0.06}  & \textbf{0.555±0.073} & 0.605±0.08           & 0.626±0.092          & 0.719±0.06           & 0.773±0.064          & 0.834±0.064          & 0.911±0.028          \\ \hline
\multirow{4}{*}{DSADS}   & Baseline                                                                                           & 0.605±0.074          & 0.647±0.059          & 0.617±0.059          & 0.658±0.048          & 0.624±0.066          & 0.664±0.059          & 0.628±0.075          & 0.668±0.065          \\
                         & Masked Reconstruction                                                                              & 0.599±0.067          & 0.642±0.062          & 0.607±0.074          & 0.646±0.059          & 0.625±0.08           & 0.663±0.065          & 0.633±0.073          & 0.673±0.061          \\
                         & Multitask                                                                                          & 0.598±0.064          & 0.639±0.056          & 0.599±0.073          & 0.64±0.063           & 0.621±0.069          & 0.659±0.062          & 0.654±0.065          & 0.693±0.058          \\
                         & PIM                                                                                  & \textbf{0.616±0.079} & \textbf{0.659±0.067} & \textbf{0.642±0.055} & \textbf{0.682±0.049} & \textbf{0.658±0.075} & \textbf{0.695±0.064} & \textbf{0.656±0.076} & \textbf{0.697±0.065}
\end{tabular}
}
\vspace{-5mm}
\label{tab:comparison_results2}
\end{table*}

\subsection{Comparison methods and baseline}

The baseline model consists of the encoder and classification head defined above and is fully supervised, trained from scratch.
The multitask implementation was based on \cite{saeed2019multi}, using the same encoder. The encoder architecture matches that of the proposed method, while each task head utilizes two linear layers with a ReLU activation in between. During training, multitask optimization was performed using binary cross-entropy loss. Following previous work \cite{yuan2024self}, the pretraining tasks included horizontal flipping, axis permutation, and time warping. 
The masked reconstruction was implemented as follows \cite{haresamudram2020masked}, using the same encoder and a reconstruction head consisting of three layers of linear-ReLU-Dropout-BatchNormalization. The network reconstructs the disturbed original signal from 10\% of the masked input, optimized using the mean squared error (MSE) loss.

In short, all methods use the same classifier architecture, consisting of a fully-connected layer with output size $256$, followed by ReLU activation, and finally, another fully-connected layer whose output size will be the same as the number of classes for each dataset.

\subsection{Comparison results}

To evaluate the effectiveness of the proposed physics-informed pre-training approach, we compare the proposed method with the state-of-the-art methods, including multi-task learning with data augmentations \cite{saeed2019multi} and mask reconstruction \cite{haresamudram2020masked}. These methods were evaluated by varying the amount of available labeled data per activity class. The results are summarized in \cref{tab:comparison_results1} and \cref{tab:comparison_results2}, which report the activity recognition performance on four benchmark datasets. \cref{tab:comparison_results1} presents results for scenarios with limited labeled data per class from 2 to 8 samples, while \cref{tab:comparison_results2} shows results for settings with more extensive labeling from 16 samples to the entire dataset.

In low-labeled data scenarios in \cref{tab:comparison_results1}, the proposed PIM approach consistently outperforms both the multi-task and masked reconstruction baselines across all datasets. Here we see the proposed method maintaining higher recognition performance with limited labeled data. This consistent advantage highlights the robustness of our method in scenarios with minimal labeled data. The physics-informed pre-training seems to better capture essential activity-specific features, possibly due to its attention to inherent physical constraints that improve generalization from limited samples. This observation indicates that our approach is particularly effective in settings where labeling resources are constrained, enabling improved accuracy without relying on extensive labeled data.

When moving to the higher labeled data setting in \cref{tab:comparison_results2}, we observe that the performance gap between the pretraining methods and the baseline narrows across all methods. With more labeled samples, the benefit of pretraining diminishes across all methods. The proposed method still performs slightly better than others, but the improvements from pretraining are generally modest. This trend demonstrates that while pretraining is crucial in low-data regimes, its relative impact diminishes as labeled data availability increases. Furthermore, in the higher data regime, the improvements offered by the proposed PIM approach are less impactful, suggesting that the pretraining gains plateau when the network is sufficiently trained on extensive labeled data.

The proposed PIM approach proves especially valuable when labeled data is limited, providing substantial gains in performance over other methods. As labeled data volume increases, the added complexity of pretraining methods becomes less critical, with traditional supervised learning approaches beginning to suffice. However, for applications where labeled data is expensive or challenging to obtain, our approach remains advantageous.

\subsection{Ablation Study}

\newcommand{\mr}[1]{\begin{tabular}{@{}c@{}} #1 \end{tabular}}
\begin{table}[!t]
\caption{Ablation study on including different types of pseudo-labels in the loss function.}
\centering
\resizebox{\linewidth}{!}{%
\begin{tabular}{l|l|ll|ll}
                         &                       & \multicolumn{4}{c}{Fine-tuning}                                                   \\
\multirow{2}{*}{Dataset} & \multirow{2}{*}{Loss} & \multicolumn{2}{c}{8 samples per class} & \multicolumn{2}{c}{10\% of samples per class}   \\
                         &                                     & F1                 & ACC               & F1               & ACC             \\
                         \hline
\multirow{4}{*}{PAMAP2}  & Baseline                            & 0.44±0.187         & 0.511±0.17        & 0.562±0.17       & 0.633±0.131     \\
                         & Angle                               & 0.523±0.175        & 0.604±0.144       & 0.592±0.181      & 0.67±0.144      \\
                         & Motion                              & \textbf{0.549±0.169}        & \textbf{0.627±0.117}       & \textbf{0.627±0.198}      & \textbf{0.695±0.146}     \\
                         & Angle+Motion                         & 0.541±0.176        & 0.613±0.131        & 0.606±0.191      & 0.675±0.149     \\
                         \hline
\multirow{8}{*}{WEAR}    & Baseline                            & 0.387±0.128        & 0.366±0.124       & 0.609±0.184      & 0.664±0.153     \\
                         & Angle                               & 0.405±0.125        & 0.386±0.123       & 0.617±0.181      & 0.674±0.155     \\
                         & Motion                              & 0.469±0.112        & 0.43±0.107        & 0.651±0.178      & 0.692±0.147     \\
                         & Synchronous                         & 0.436±0.136        & 0.4±0.127         & 0.645±0.182      & 0.686±0.146     \\
                         & Angle+Motion                        & 0.475±0.12         & 0.442±0.114       & 0.651±0.172      & 0.697±0.139     \\
                         & Angle+Synchronous                   & 0.446±0.138        & 0.424±0.132       & 0.642±0.178      & 0.687±0.147     \\
                         & Motion+Synchronous                  & \textbf{0.475±0.122}        & \textbf{0.438±0.11}        & 0.656±0.173      & 0.697±0.142     \\
                         & \mr{Angle+Motion\\+Synchronous}     & 0.452±0.109        & 0.423±0.108       & \textbf{0.661±0.172}      & \textbf{0.708±0.14}      \\
                         \hline

\multirow{8}{*}{DSADS}   & Baseline                            & 0.535±0.073        & 0.585±0.062       & 0.581±0.092      & 0.626±0.077     \\
                         & Angle                               & 0.523±0.094        & 0.576±0.087       & 0.584±0.069      & 0.632±0.061     \\
                         & Motion                              & 0.585±0.081        & 0.631±0.068       & 0.602±0.078      & 0.642±0.061     \\
                         & Synchronous       & \textbf{0.609±0.08}         & \textbf{0.652±0.066}       & 0.618±0.05       & 0.663±0.045     \\
                         & Angle+Motion                        & 0.581±0.078        & 0.625±0.068       & 0.584±0.081      & 0.629±0.065     \\
                         & Angle+Synchronous                   & 0.598±0.083        & 0.64±0.071        & 0.608±0.072      & 0.649±0.062     \\
                         & Motion+Synchronous                  & 0.563±0.089        & 0.607±0.08        & \textbf{0.62±0.08}        & \textbf{0.659±0.069}     \\
                         & \mr{Angle+Motion\\+Synchronous}     & 0.592±0.075        & 0.636±0.064       & 0.575±0.08       & 0.616±0.071    
\end{tabular}
}
\vspace{-5mm}
\label{tab:results:ablation:loss}
\end{table}

Since our PIM method pre-trains on different types of pseudo-labels, we conducted an ablation study to highlight how each type contributes to the obtained classification results, be it alone or in combination with others. In \cref{tab:results:ablation:loss} we can see the results for this ablation. In other words, for each dataset, we pre-trained with different terms of our loss function \cref{eq:loss}. For this test, we evaluated the results under two scenarios: with 8 labeled samples per class and with 10\% of samples per class for fine-tuning. 

As we can see in \cref{tab:results:ablation:loss}, for the PAMAP2 and WEAR datasets, all versions of our loss function improved on the baseline of fully supervised training without pre-training. This indicates that PIM’s pseudo-label-based tasks generally capture useful representations, improving the performance of HAR even when fine-tuning data is limited. 

For the DSADS dataset, however, angle-based features alone did not improve on the baseline when fine-tuning with only 8 samples per class, and only marginally improved on it when 10\% of samples per class were available for fine-tuning. This limited effect may suggest that angle features are less informative for this dataset’s specific activity patterns when labeled data is lacking. Nonetheless, other pseudo-label components, particularly synchronicity-based features, consistently outperformed the baseline, making the largest positive impact across DSADS configurations. Interestingly, while our angle-based features improve on the baseline, they are part of the best approach only for the WEAR dataset with 10\% of samples per class for fine-tuning. This suggests that angle features may capture trends relevant to certain dataset characteristics, particularly when combined with other pseudo-label types.

\subsection{Limitations and Future Work}
While our PIM method achieves impressive improvements in classification performance, especially with very limited labeled data, several limitations still remain. Some of our defined physical quantities, particularly those related to synchronicity, require multiple sensor locations. This dependency restricts our approach in real-world situations where users may be wearing only a single device, such as a smartphone or smartwatch. Moreover, we have not investigated our method for this single-device case, which is compatible with our angle and motion pseudo-labels. We also have yet to explore pre-training on pseudo-labels from multiple devices and transferring this knowledge to a network that uses a single one. Future work could explore pre-training on multi-sensor pseudo-labels and transferring that knowledge to single-device models, potentially through techniques like knowledge distillation or shared encoders across sensor modalities.

Another limitation of the current approach is its handling of unbalanced datasets.
Future work also includes directly dealing with unbalanced datasets, which could be done by re-weighting samples or using a custom sampling approach. As we have seen in our experiments on the MM-FIT dataset, our method can help even in unbalanced datasets, but as it is now it fails to improve on the baseline as more than 32 examples per class are available. Future work also includes evaluating different encoder architectures, as well as pre-training across datasets, which could significantly increase the amount of pre-training data.

\section{Conclusion}
\label{sec:conclusion}
In this paper, we introduced a novel physics-inspired pre-training (PIM) method for HAR, leveraging pseudo-labels based on physical quantities to improve the effectiveness of sensor data pre-training. By incorporating meaningful features such as symmetry of the human body, angular information of moving devices, and speed of motion, our PIM approach captures relevant physical relationships inherent in human activities, which are often overlooked in traditional data-driven pre-training methods. Through comprehensive experiments across multiple benchmark datasets, we demonstrated that PIM consistently outperforms state-of-the-art approaches, particularly in situations with low data availability of labeled examples, where the need for efficient generalization from limited labeled samples is critical. Our ablation study further highlighted the contributions of each physical pseudo-label, showing how different features uniquely support model performance across various datasets and sample availability levels.

While our results demonstrate the potential of PIM, the method still has limitations, such as its reliance on multiple sensors for certain pseudo-labels and challenges in handling unbalanced data in high-sample scenarios. Future work will explore adaptations for single-device settings, as well as enhanced strategies for handling imbalanced datasets, cross-dataset pre-training, and expanded architecture options.

\section*{Acknowledgment}
The research reported in this article was supported by the Carl Zeiss Stiftung, Germany, in the Sustainable Embedded AI project (P2021-02-009) and by the European Union in the SustainML project (Grant agreement ID: 101070408).

\bibliographystyle{IEEEtran}
\bibliography{ref}

\end{document}